\documentclass{article}

\usepackage[numbers]{natbib}

     \usepackage[final]{arxiv_version}

\usepackage[utf8]{inputenc} 
\usepackage[T1]{fontenc}    
\usepackage{hyperref}       
\usepackage{url}            
\usepackage{booktabs}       
\usepackage{amsfonts}       
\usepackage{nicefrac}       
\usepackage{microtype}      
\usepackage{xcolor}
\usepackage[pdftex]{graphicx}

\title{Estimating Skin Tone and Effects on Classification Performance in Dermatology Datasets }

\author{%
  Newton M. Kinyanjui,$^{1,4}$ Timothy Odonga,$^{1,4}$ Celia Cintas,$^1$ Noel C. F. Codella,$^2$ \\\bf{Rameswar Panda,$^3$ Prasanna Sattigeri,$^2$ and Kush R. Varshney$^{1,2}$} \\
  IBM Research, $^1$Nairobi, Kenya, $^2$Yorktown Heights, NY, USA, $^3$Cambridge, MA, USA\\
  $^4$Carnegie Mellon University Africa, Kigali, Rwanda\\
}

\begin{document}

\maketitle

\begin{abstract}
    Recent advances in computer vision and deep learning have led to breakthroughs in the development of automated skin image analysis. In particular, skin cancer classification models have achieved performance higher than trained expert dermatologists. However, no attempt has been made to evaluate the consistency in performance of machine learning models across populations with varying skin tones. In this paper, we present an approach to estimate skin tone in benchmark skin disease datasets, and investigate whether model performance is dependent on this measure. Specifically, we use individual typology angle (ITA) to approximate skin tone in dermatology datasets. We look at the distribution of ITA values to better understand skin color representation in two benchmark datasets: 1) the ISIC 2018 Challenge dataset, a collection of dermoscopic images of skin lesions for the detection of skin cancer, and 2) the SD-198 dataset, a collection of clinical images capturing a wide variety of skin diseases. To estimate ITA, we first develop segmentation models to isolate non-diseased areas of skin. We find that the majority of the data in the the two datasets have ITA values between 34.5$^\circ$ and 48$^\circ$, which are associated with lighter skin, and is consistent with under-representation of darker skinned populations in these datasets. We also find no measurable correlation between performance of machine learning model and ITA values, though more comprehensive data is needed for further validation. 
\end{abstract}

\section{Introduction}
As machine learning is becoming more frequently applied to support decisions that affect the lives of many, understanding how to accurately measure latent dataset characteristics and demographic representation is gaining increasing interest to prevent the potential negative consequences of dataset imbalances \cite{rotemberg2019role}, henceforce referred to as ``dataset bias''. Dataset bias is a critical issue because it is one of the causes of machine learning-based systems placing certain groups of people at a systematic disadvantage \cite{barocas2016big}. Recognition and mitigation of unwanted bias throughout the machine learning pipeline is necessary in order to build machine learning systems that are trusted in their eventual domains of deployment \cite{bellamy}.

Skin diseases continue to bear significant negative impacts on human health globally. A 2017 study found that skin diseases contribute 1.79$\%$ to the global burden of disease measured in disability-adjusted life years \cite{karimkhani2017global}. Skin diseases such as skin cancer account for about 7$\%$ of new cancer cases globally \cite{bray2018global}, with a cost to the US healthcare system that exceeded 8 billion in 2011 \cite{cdc2014}. Within skin cancer cases, there is evidence of some outcome disparities with respect to ethnicity: although people of color are roughly 20 to 30 times less likely to develop melanoma than lighter skinned individuals, for certain melanoma sub-types they have been found to have lower \cite{marchetti2015screening,wu2011racial,mahendraraj2017} or higher \cite{mahendraraj2017} survival rates.  Some studies have found that for people of color, the diagnosis of skin cancer may occur at a more advanced stage, leading to lower rates of survival and poorer outcomes \cite{gohara2015skin,kundu2013dermatologic}. However, increased screening using currently available diagnostic resources also carries risks and can lead to significant harms, such as unnecessary surgeries, disfigurement, disability, morbidity, and over-diagnosis  \cite{marchetti2015screening}. 

Computer vision has been studied in the context of dermatology image analysis for decades, and several review articles have been written \cite{STOECKER1992145,KOROTKOV201269,ibmbookchapter}. The success of deep learning models has led to studies applying the technology to dermatological use cases \cite{8627921,8660594}. Models using convolutional neural networks (CNNs) have been applied to problems such as skin cancer diagnosis and were found to outperform trained dermatologists in controlled settings and datasets \cite{Codella2016,haenssle2018man,naturepaper}. However, as most of the publicly available datasets of skin images come from lighter skinned populations, due to the extreme disparities in disease prevalence, there are concerns about how to best collect data, train, and evaluate models for darker skinned populations \cite{adamson2018machine,rotemberg2019role}. Also, because of the significant risks of harm from over-diagnosis with increased screening in low-risk dark skin populations, there is a need to better discriminate between life-threatening and stable presentations of disease \cite{marchetti2015screening,rotemberg2019role}.

In this paper, we work towards answering questions around quantifying distributions according to estimated skin tone in datasets where this information is currently unavailable, and measuring downstream effects on classifier performance. Specifically, our contributions are as follows:

\begin{itemize}
    \item We propose a pipeline to automatically estimate skin tone for images in two public benchmark skin disease datasets using the individual typology angle (ITA), which has been used previously as a measure of skin tone in absence of manually curated information \cite{Merler2019}. 
    \item We have generated both a subset of manually delineated ground truth segmentation masks, as well as automatically generated masks, for non-diseased skin in both public benchmarks.
    \item We quantitatively confirm that the two benchmark skin disease datasets under-represent ITA values that would generally be correlated with darker skin populations. 
    \item No correlation between performance of machine learning model and ITA values are measureable at this time, though more data is needed for conclusive results. 
\end{itemize}

We hope that this work will encourage additional work to help focus resources toward further data collection to fill existing gaps, and improve classifier evaluations for comprehensive assessments across demographics.

\section{Related work}
Over recent years, there have been significant advances in automated skin lesion analysis, with hundreds of deep learning models implemented for skin cancer diagnosis. Much of this work has been enabled and supported by the International Skin Imaging Collaboration (ISIC) \cite{ISIC2018,Codella2019,Tschandl2018}, which has organized a public repository of annotated dermoscopic images, and hosted 4 consecutive years of public challenge benchmarks. In 2016, the first work demonstrating classification accuracy higher than the average of expert dermatologists was described \cite{Codella2016}, employing an ensemble of methods that included hand-coded feature extraction, sparse coding methods, support vector machines, CNNs for skin lesion classification, and fully convolutional networks for skin lesion segmentation. Other models have also been implemented by researchers, such as a computationally efficient skin lesion classification model that uses the MobileNet architecture implemented by \cite{chaturvedi2019skin}, and an Inception architecture trained on a large dataset of over 100,000 images \cite{naturepaper}.

Outside of dermatology, there has been work on evaluating fairness in computer vision with respect to skin type. Recent studies evaluated bias in automated facial analysis algorithms and datasets with respect to phenotypic groups \cite{buolamwini2018gender,raji2019actionable}. They found substantial disparities in the accuracy of classifying darker females, lighter females, darker males, and lighter males in gender classification systems with darker-skinned females as the most misclassified group \cite{buolamwini2018gender,raji2019actionable}. A related study on gender classification systems tested the stability to skin type of a gender classification system by varying the skin type of a face keeping all other features fixed, statistically showing that the effect of skin type on classification outcome was minimal \cite{muthukumar2019color}. The study revealed that well-performing gender classification systems are already invariant to skin type and thus the skin type by itself has a minimal effect on classification disparities \cite{muthukumar2019color}. Another study investigated equitable predictive performance in state-of-the-art object detection systems on pedestrians with different skin types, finding higher precision on lower Fitzpatrick skin types (lighter skin) than higher skin types (darker skin) \cite{wilson2019predictive} .

Most of the work in automated skin lesion analysis has focused on creating benchmark models. Our work focuses on the analysis of some of the benchmark models with respect to different skin tones that could represent different ethnicities. The work in this paper builds on the work done from previous ISIC challenges by investigating the accuracy of a skin lesion classification model over a range of skin tone categories.

\section{Datasets}
Public benchmark datasets, in addition to fostering direct comparisons among various algorithms to facilitate advancement in terms of classification performance, are also capable of supporting detailed analysis of that performance with respect to various characteristics of the dataset \cite{rotemberg2019role}.  Therefore, we focus our analysis on two of the most widely used dermatology datasets in the computer vision literature: the ISIC 2018 Challenge dataset the SD-198 dataset.

\paragraph{ISIC2018:} This dataset consists of a collection of dermoscopic images, separated into datasets for 3 tasks of image segmentation (Task 1), clinical feature detection (Task 2), and disease classification (Task 3). Dermoscopic images are acquired through a digital dermatoscope, with relatively low levels of noise and consistent background illumination. The training dataset for Task 3 is the largest among the tasks and used in this work. It consists of 10,015 dermoscopic images publicly available through the ISIC archive \cite{Codella2019,Tschandl2018}, falling into one of 7 skin diseases; melanoma, melanocytic nevus, basal cell carcinoma, actinic keratosis, benign keratosis, dermatofibroma and vascular lesion.

\paragraph{SD-198:} The SD-198 dataset, a large scale benchmark dataset for visual recognition of skin diseases, contains 6,548 clinical images from 198 skin disease classes, varying according to scale, color, shape and structure, and consists of images downloaded from DermQuest, an online medical resource for dermatologists \cite{sun2016benchmark}. Clinical images are collected via various devices, most of which are digital cameras and mobile phones \cite{sun2016benchmark}. Higher levels of noise and varying illumination in clinical images makes segmentation more challenging for this dataset in comparison to the dermoscopic images in ISIC2018.  Each image in this dataset has been labeled by experts to denote the disease class. 

In this work, we preprocess the SD-198 dataset to exclude classes containing images with no observable non-diseased skin. Eventually 136 disease classes are retained, and henceforth the pre-processed dataset is referred to as ``SD-136''. Some of the classes excluded from the SD-136 dataset include classes of lesions inside the mouth, such as fibroma, geographic tongue, and stomatitis. Other diseases such as arsenical keratosis, pustular psoriasis, and mal perforans contain images of lesions on palms and soles of the feet from which it is difficult to determine the individual's skin tone. Other disease classes such as stasis ulcer and eccrine poroma contained images of severely scarred skin from which it is visually impossible to differentiate non-diseased and diseased skin. This preprocessing step was done manually and eventually 4,467 images were retained from the original 6,548 images.

Since there are no existing ground truth segmentation masks for the SD-136 dataset, we manually segmented a subset of 343 images using the ImageJ image processing software. We were particularly interested in segmenting regions with non-diseased skin from other regions of the image containing diseased skin, shadows, and other artifacts. We used these ground-truth masks in training the segmentation model for the SD-136 dataset.

\section{Methods}
The work presented is a two-step approach. First, we quantify the skin tone categories represented in benchmark datasets. For this, we segment skin images to extract non-diseased background regions, and use a metric to characterize the skin tone of that region. Second, we evaluate the classification performance across different skin tone categories. 

\begin{figure}
    \centering
    \includegraphics[width=0.8\linewidth]{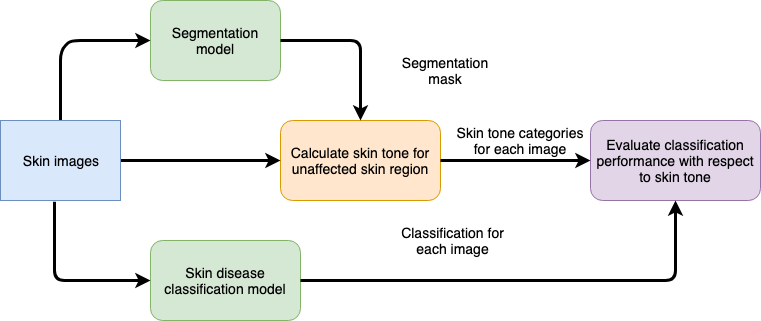}
    \caption{Block diagram of methodology.}
    \label{fig:blockdiagram}
\end{figure}

This is achieved through a series of steps, summarized in Fig.~\ref{fig:blockdiagram}. First, we train a model to segment skin disease images to obtain the non-diseased skin in the image. Second, we select and compute the metric to stratify the non-diseased skin into a skin tone category. After that, a classification model is trained to classify skin images into one of the skin diseases in the dataset. Finally the performance of the classifier on samples in each skin tone category is evaluated. 
In the following subsections we will explain each step in detail.

\subsection{Quantification of representation of skin tone categories in benchmark datasets}

\subsubsection{Skin lesion segmentation}
Segmentation of the skin lesion from the non-diseased skin is done using a Mask R-CNN model~\cite{he2017mask}. Mask R-CNN was selected because it was one of the top performing skin lesion boundary segmentation models in the 2018 ISIC challenge \cite{ISIC2018} and also because it has been shown to be highly effective and efficient in performing semantic segmentation \cite{johnson2018adapting}. 

To obtain a segmentation model for the ISIC2018 dataset, a Mask R-CNN pretrained on the COCO dataset is finetuned with the lesion boundary segmentation data from ISIC 2018 Challenge Task 1. The data is split into training and validation data using 90$\%$ to 10$\%$ train validation split. The images are resized to $600\times 450$ pixels to correspond to the size used for classification. Random horizontal flips are done on the images for data augmentation during training. The segmentation model for the ISIC2018 dataset is trained for $25$ epochs. This model is used to predict segmentation masks for the all the classification data. An additional step of thresholding based on contour extraction on the predicted masks is implemented using the OpenCV library to improve the segmentation masks.

The segmentation model from the ISIC2018 dataset is finetuned with the 343 images from SD-136 having ground truth segmentation masks. The data is split into training and validation data using 90$\%$ to 10$\%$ train validation split. The images are resized to $450 \times 450$ pixels. Random horizontal flips are done on the images for data augmentation during training. The segmentation model for the SD-136 dataset is trained for 50 epochs. This model is used to predict segmentation masks for the all the classification data in SD-136. To get binary masks, a thresholding step is further done on the grayscale masks predicted from the model.

The quality of segmentation is evaluated using the accuracy and the false negative rate of the predicted segmentation masks with respect to the ground-truth segmentation masks. The false negative rate is considered because it is more dangerous to wrongly classify a diseased region as non-diseased in our analysis. To further evaluate the quality of segmentation, mean absolute error is computed between the ITA estimates from ground truth masks and ITA estimates from predicted masks.

\subsubsection{Skin tone categorization}
With the segmentation masks obtained from the previous step, we are able to obtain pixels in the non-diseased region for each image. We use these pixels to categorize the skin tone of each sample. 

There is no universal method for characterizing skin type or skin tone among dermatologists.  The Fitzpatrick skin type, used in \cite{buolamwini2018gender,raji2019actionable,muthukumar2019color}, is a dermatologist's determination of a person's risk of sunburn.  It is by definition, however, a subjective human determination \cite{EilersBGBGNKR2013}.  In contrast, the melanin index is measured objectively via reflectance spectrophotometry, and has a strong correlation with the Fitzpatrick type and is useful in assigning it \cite{WilkesWPR2015}.  The metric we use to quantify skin tone in this work is the individual typology angle (in degrees) because it has strong (anti-)correlation to the melanin index \cite{WilkesWPR2015}, and can be simply computed from images, making it a practical method for categorizing skin color \cite{Merler2019}. 

The pixels from the non-diseased region are converted to CIELab-space to obtain the L and b values. L quantifies the luminance of each pixel and b quantifies the amount of yellow in each pixel. The mean of L and b values are computed. The mean ITA value for the pixels in the non-diseased region is calculated using \cite{Merler2019}:
    $$ \rm ITA = \arctan{\bigg(\frac{L-50}{b}\bigg)}\times \frac{180^{\circ}}{\pi} . $$
To deal with outliers, L and b values applied to the formula are kept within one standard deviation. For each image the mean ITA value is computed from the ITAs obtained.

The ITA values computed for the images are binned in a scheme similar to that employed by \cite{casale2015extreme}, which used 5 skin tone categories, namely: Very Light, Light, Intermediate, Tanned, and Dark. The Light, Intermediate, and Tanned categories are further subdivided into two equal ranges each for better analysis, thus giving a total of 8 ITA categories. The scheme used to categorize skin tones is summarized in Table \ref{table:categorizationscheme}. 

\begin{table}
  \caption{Skin tone categorization scheme.}
  \label{table:categorizationscheme}
  \centering
  \begin{tabular}{lll}
    \toprule
    ITA Range     & Skin Tone Category     & Abbreviation \\
    \midrule
    ITA $>55^{\circ}$ & Very Light & very\_lt  \\
    $48^{\circ}<$ ITA $\leq 55^{\circ}$ & Light 2 & lt2\\
    $41^{\circ}<$ ITA $\leq 48^{\circ}$ & Light 1 & lt1\\
    $34.5^{\circ}<$ ITA $\leq 41^{\circ}$ & Intermediate 2 & int2\\
    $28^{\circ}<$ ITA $\leq 34.5^{\circ}$ & Intermediate 1 & int1\\
    $19^{\circ}<$ ITA $\leq 28^{\circ}$ & Tanned 2 & tan2\\
    $10^{\circ}<$ ITA $\leq 19^{\circ}$ & Tanned 1 & tan1\\
    ITA $\leq10^{\circ}$ & Dark & dark \\
    \bottomrule
  \end{tabular}
\end{table}

\subsection{Evaluation of classification performance across different skin tone categories.}
\subsubsection{Skin disease classification}
To perform skin disease classification, a Densenet201 model pretrained on the ImageNet dataset is finetuned using our training data. The Densenet201 model is chosen because it was one of the best performing single models for lesion classification in the ISIC 2018 challenge \cite{ISIC2018}. 

The ISIC2018 dataset comprises 10,015 images which fall under one of 7 disease classes. The images are maintained at $600\times450$ pixels. Additional transformations such as random horizontal flipping are applied to augment the data. The samples in each batch are normalized using the mean and standard deviation computed on all samples in the dataset to ensure fast convergence during training. The data is split into training and validation data using an 80$\%$ to 20$\%$ train validation split. A weighted cross entropy loss function and an Adam optimizer are used for training. The weights for the loss function are obtained from the inverse of each disease class frequency. This loss function is chosen because it accounts for class imbalance in the dataset.

The SD-136 dataset comprises 4467 images which fall under one of 136 disease classes. The images are first resized to $450\times 450$ pixels and then center cropped to $360\times 360$ pixels. Additional random transformations including random horizontal flipping and random rotation by a degree selected between $-90^{\circ}$ and $90^{\circ}$ are applied to augment the data during training. Similarly, samples in each batch are normalized using the mean and standard deviation. The data is split into training and validation data using an $80\%$ to $20\%$ train validation split. A weighted cross entropy loss function and an Adam optimizer are used for training.

During training, the early layers up to and including the first Dense block are frozen and all the successive layers have their weights updated. Each classification model is trained for 300 epochs with a patience of 100 epochs at which early stopping would be applied to prevent overfitting. 

The performance of the skin disease classification models is measured using the normalized multi-class accuracy metric used in the 2018 ISIC challenge \cite{Codella2019}.

\subsubsection{Performance evaluation for each skin tone category}
The goal is to investigate how classification models perform across the spectrum of skin tones present in the data. The motivation is to investigate if there is bias in performance of the models with respect to skin tone. The accuracy of the classification model is computed for each skin tone category in the validation data.

\section{Results}
\begin{figure*}
    \centering
         \includegraphics[width=\linewidth]{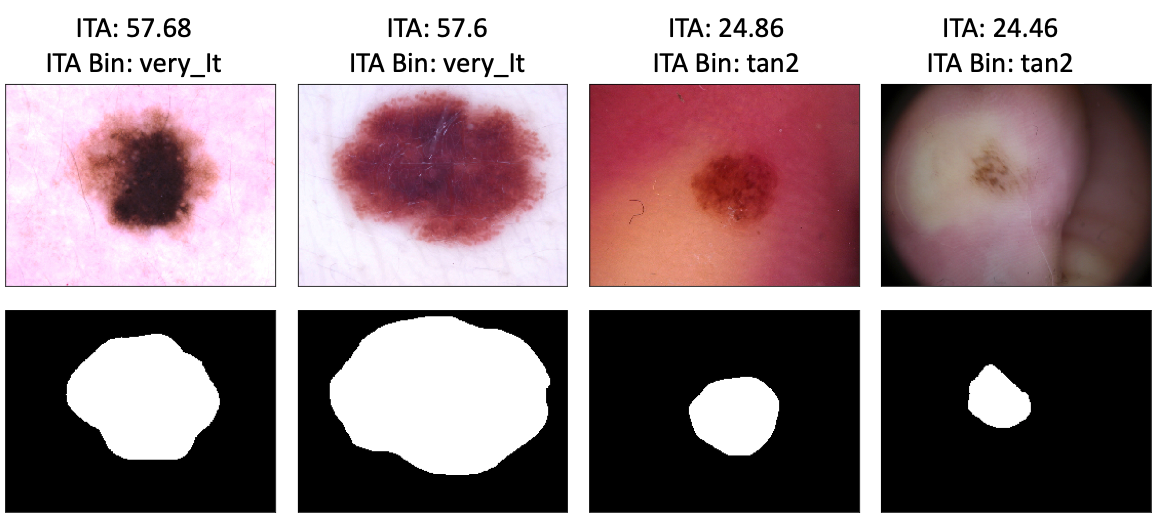}
          \caption{Sample images from ISIC2018 dataset (top row) and corresponding masks predicted by model (bottom row); ITA is computed on the non-diseased region which is colored black.}
          \label{fig:img_samples_ham}
\end{figure*}

\begin{figure*}[ht]
    \centering
            \includegraphics[width=\linewidth]{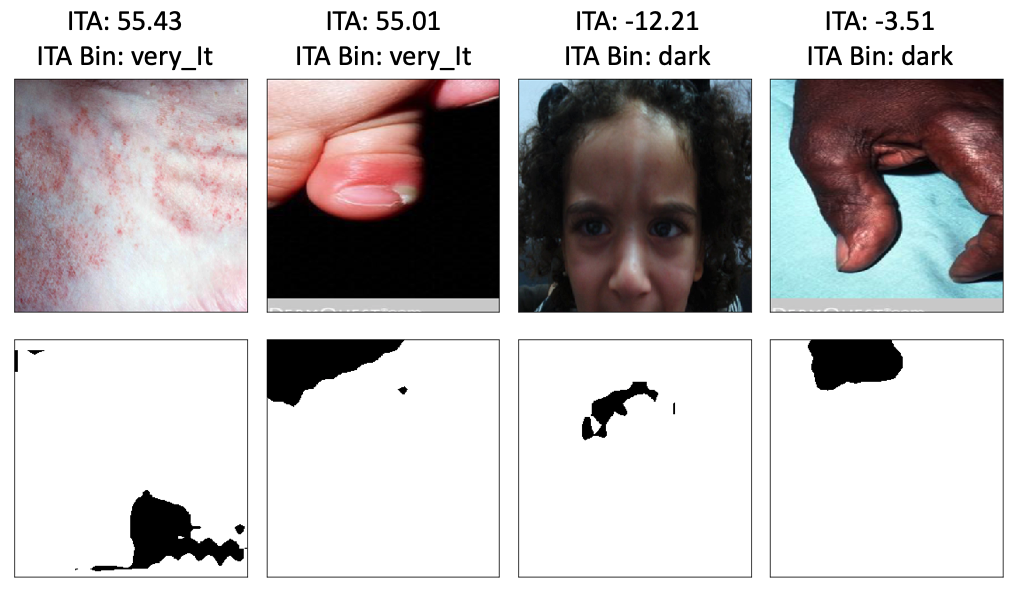}
        \caption{Sample images from SD-136 dataset (top row) and corresponding masks predicted by model (bottom row); ITA is computed on the non-diseased region which is colored black.}
        \label{fig:img_samples_sd136}        
\end{figure*}

The Mask R-CNN model used for segmentation on the ISIC2018 dataset yields an accuracy of 0.956, a false negative rate of 0.024, and a mean absolute error in ITA computation of 0.428 degrees. The segmentation model on the SD-136 dataset yield an accuracy of 0.802, a false negative rate of 0.076, and a mean absolute error in ITA computation of 3.572 degrees. These are all fairly good results and sufficient for further analysis.
Examples with segmented mask and ITA values for both datasets are shown in Fig.~\ref{fig:img_samples_ham}--\ref{fig:img_samples_sd136}.
The ITA values calculated were found to be highly correlated to the mean and median grayscale values of the pixels in the non-diseased region. Table~\ref{correlation-table} summarizes the correlation values of the ITA with mean an median grayscale values.
\begin{table}
  \caption{Correlation of ITA with mean and median grayscale}
  \label{correlation-table}
  \centering
  \begin{tabular}{lcc}
    \toprule
    Dataset     & correlation with mean & correlation with median \\
    \midrule
    ISIC2018 & 0.960  & 0.981     \\
    SD-136     & 0.912  & 0.937      \\
    \bottomrule
  \end{tabular}
\end{table}

\begin{figure*}[ht]
    \centering
    \begin{tabular}{cc}
        \includegraphics[width=0.52\linewidth]{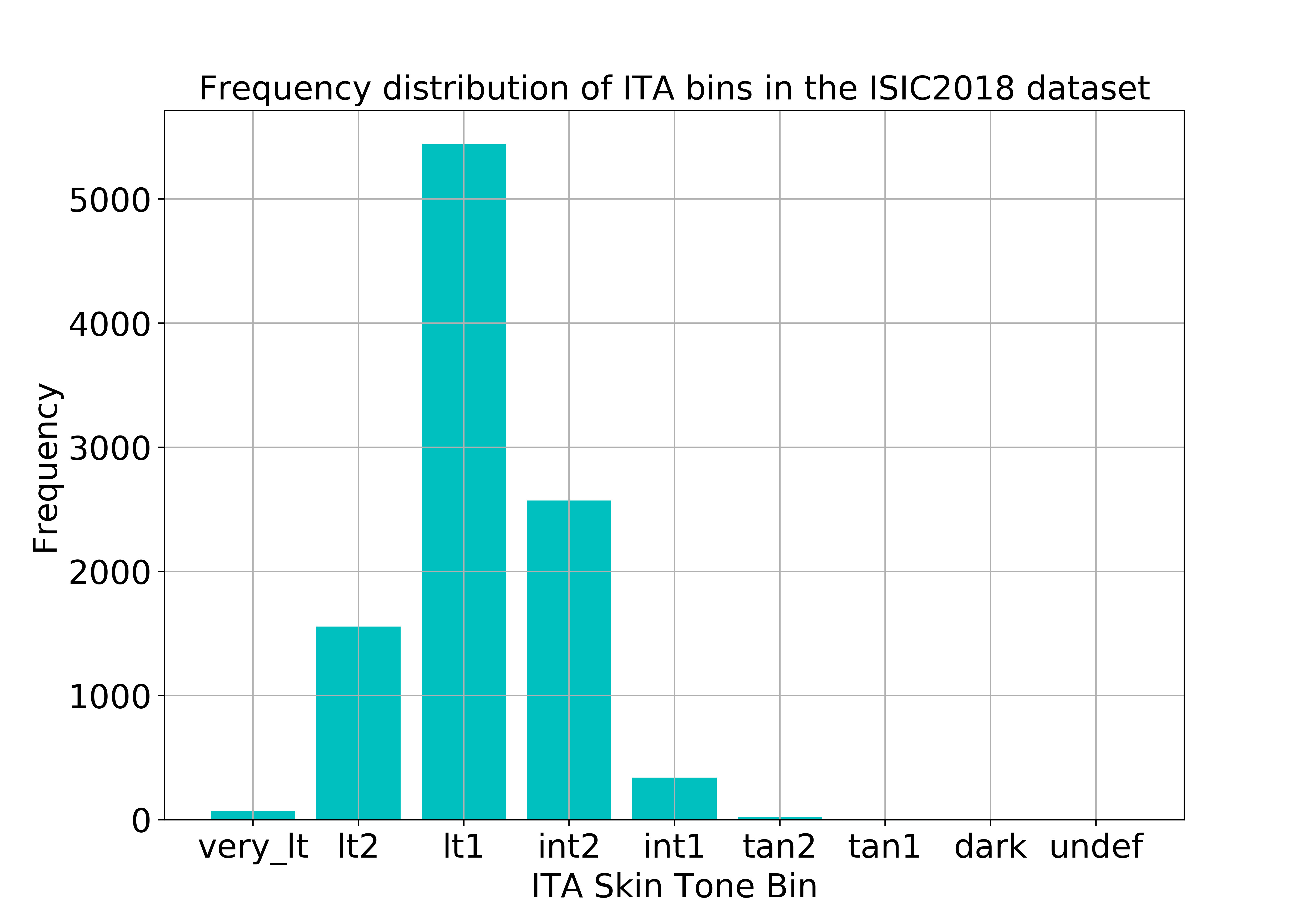} & \includegraphics[width=0.5\linewidth]{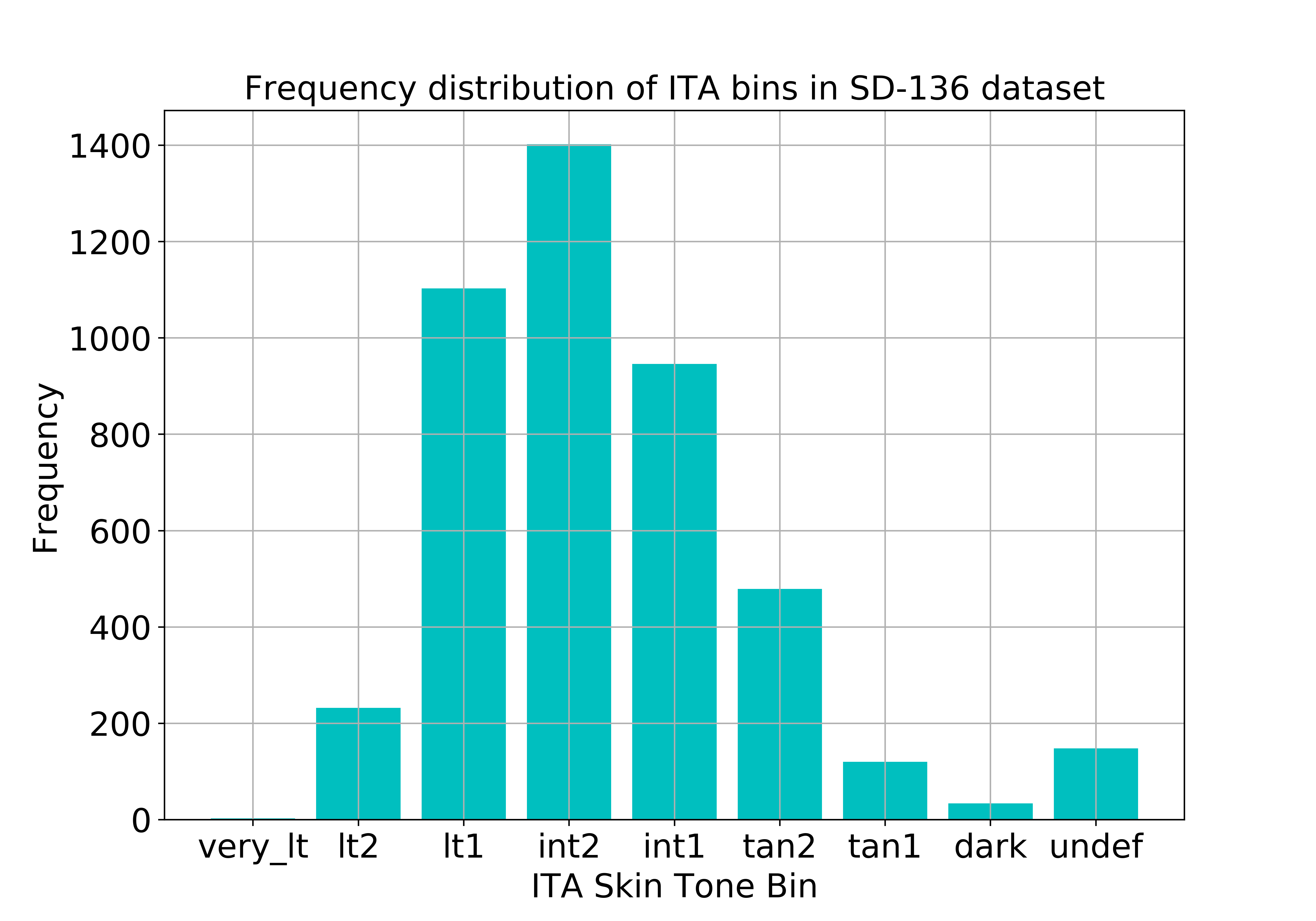} \\
        \footnotesize{(a)} & \footnotesize{(b)}
    \end{tabular}
        \caption{ITA skin tone distribution for (a) the entire ISIC2018 dataset, and (b) the entire SD-136 dataset. Most of the samples in the two datasets come from the lt2, lt1, and int2 bins. The SD-136 dataset has a broader distribution across skin tone categories with more samples in the darker region.}
        \label{fig:itadataset}
\end{figure*}

Fig.~\ref{fig:itadataset} shows the distributions of the ITA values estimated from the non-diseased skin regions of the images in the entire ISIC2018 and SD-136 datasets. Both datasets are found to predominantly lie in the Light category.  

On the ISIC2018 dataset, the Densenet201 model achieves an accuracy 0.869 and a balanced accuracy score of 0.814 on the validation partition after training approximately 140 epochs when early stopping occurred. This performance should be in the range of state-of-art systems as the values are similar to those shown on the ISIC 2018 Challenge leaderboard. Although the leaderboard results are computed on a separate held-out test set, we felt the performance of our model was good enough for subsequent analysis on dependence of ITA.

The model trained on SD-136 achieves an accuracy of 0.604 and a balanced accuracy score of 0.601. The benchmark model for SD-198 achieves an accuracy 0.52, as reported in \cite{sun2016benchmark}. However, since we dropped the number of classes from 198 to 136, we do not have a benchmark model for comparison. Nonetheless, we are confident that we have a well-performing model.

\begin{figure*}
    \centering
    \begin{tabular}{cc}                            \includegraphics[width=0.5\linewidth]{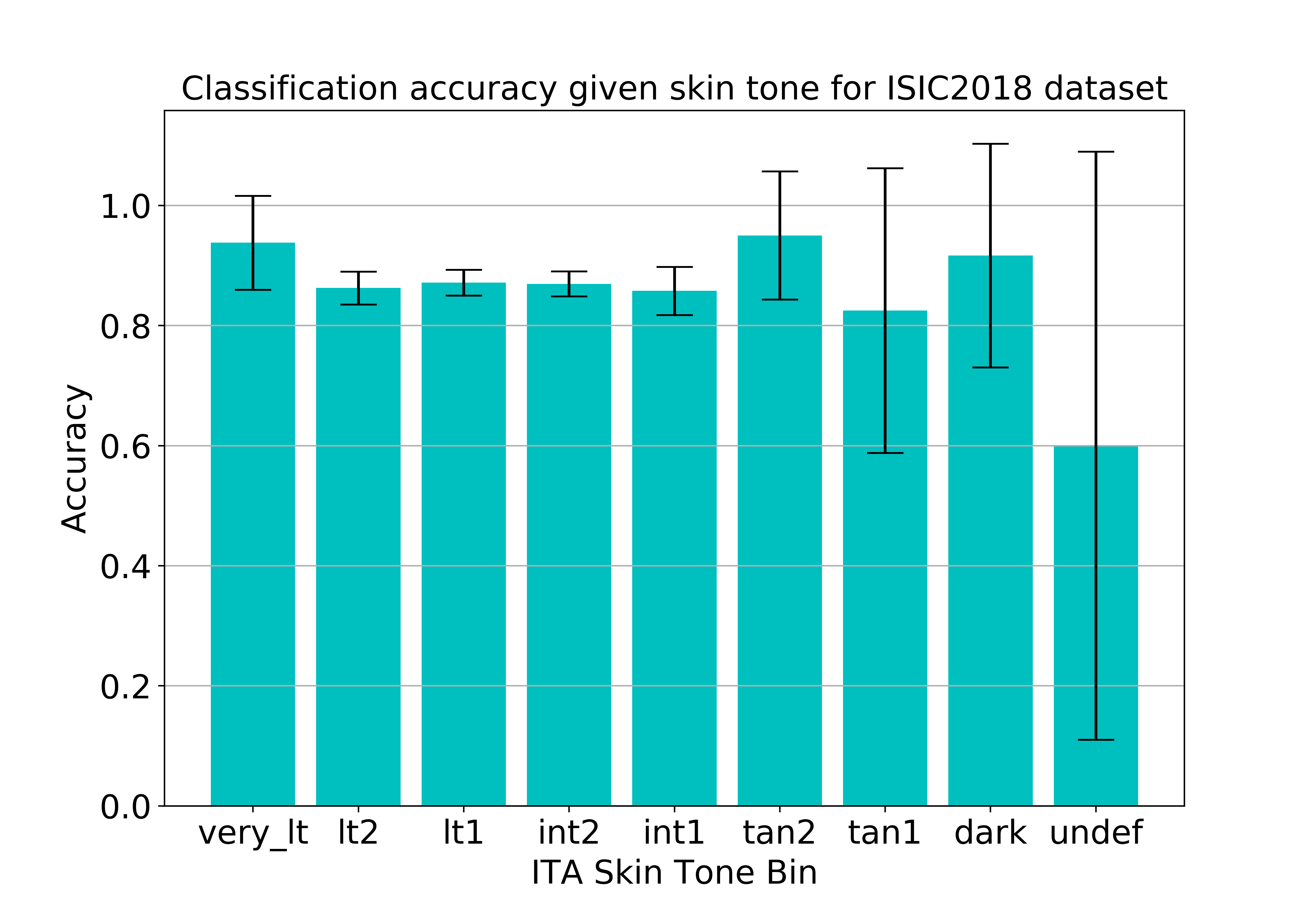} & \includegraphics[width=0.5\linewidth]{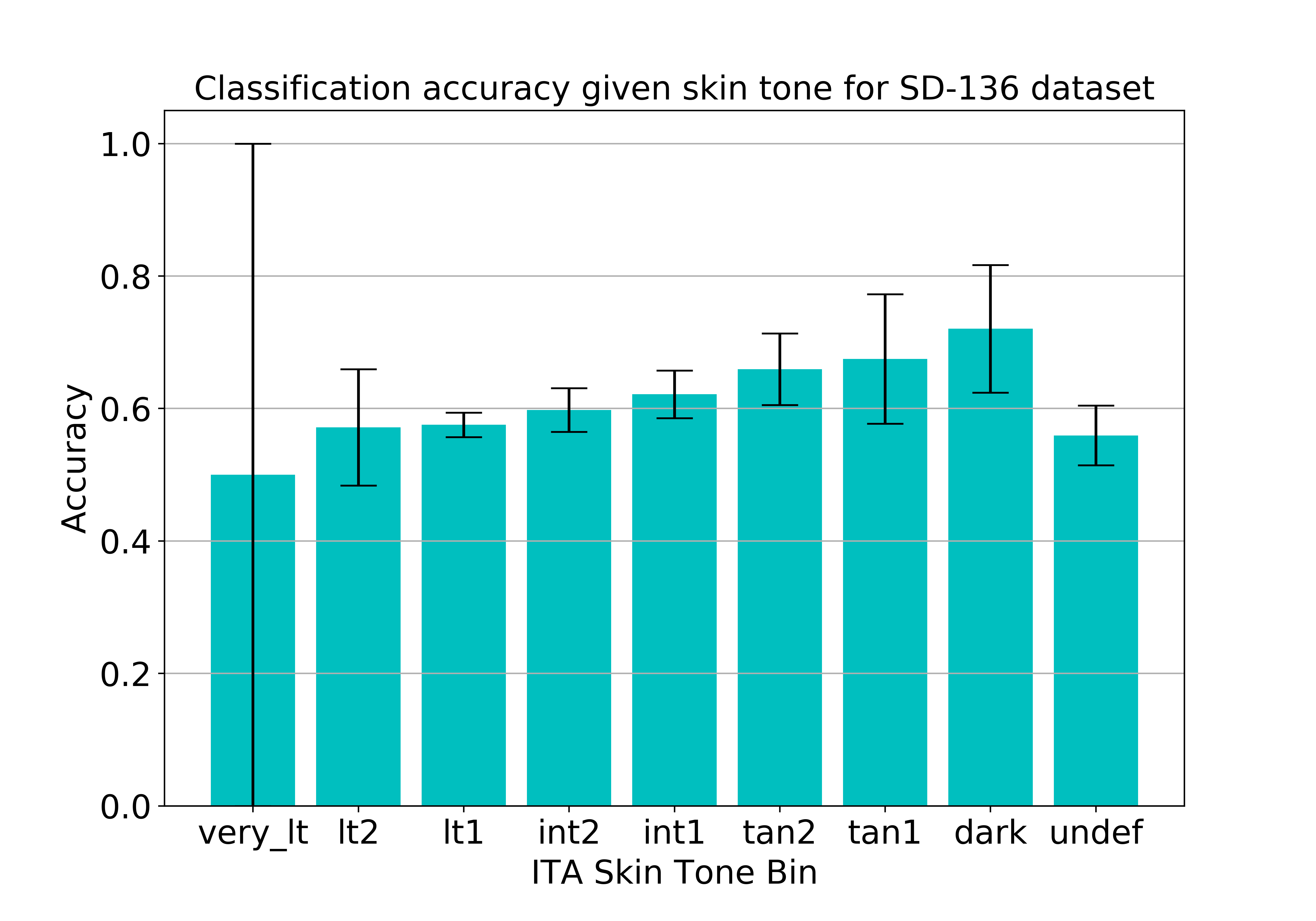} \\
    \footnotesize{(a)} & \footnotesize{(b)}
    \end{tabular}
        \caption{Classification accuracy versus ITA for (a) ISIC2018, and (b) SD-136 validation sets.}
        \label{fig:accuracy}
\end{figure*}

Importantly, on evaluating the classification performance with respect to skin tone category, our results do not show a clear trend in the performance of the model. Fig.~\ref{fig:accuracy} plots classification accuracy versus ITA for the validation set for the two datasets. The error bars indicate the standard error estimated through ten runs with random splits. Table~\ref{tab:classification-table} presents the mean accuracy score per skin tone category for the two datasets. The slope of the least squares line of best fit of the mean accuracy versus the midpoint ITA value of the bin for ISIC2018 is -0.000 (per degree) with a 95\% confidence interval of $(-0.001, 0.001)$, whereas that for SD-136 is -0.002 (per degree) with a 95\% confidence interval of $(-0.003, -0.001)$, which indicate that there are no particular trends in both datasets.  
\begin{table}
  \caption{Classification mean accuracy per skin tone. In the two datasets there is no clear trend in performance of the skin classification model across skin tone categories. The error bars indicate standard deviation across 10 random splits.}
  \label{tab:classification-table}  
  \centering
  \begin{tabular}{llllllllll}
    \toprule
    ~           & Overall &  ~      \\
    Dataset     &Accuracy  &very\_lt & lt2 & lt1 & int2 & int1 & tan2 & tan1 & dark \\
    \midrule
    ISIC2018 &  0.869 & 0.94 &   0.86  &   0.87 &    0.87  &  0.86  &   0.95 &   0.83 &    0.92\\
    SD-136  &  0.604 & 0.50 &   0.57 &    0.58  &  0.60  &   0.62 &   0.66 &    0.67 &    0.72 \\
    \bottomrule
  \end{tabular}

\end{table}

\section{Discussion and Conclusion}
In this work, we implemented an approach to measure approximate skin tone distributions in public dermatology image datasets using ITA as an estimator, and evaluated the performance of dermatology classification models with respect to the resultant ITA values. 

The distribution of ITA values across both ISIC2018 and SD-136 datasets are consistent with under-representation of darker skin tones. The results from the evaluation of the accuracy of the skin classification model for each skin tone category in the validation data shows that there is no observable trend in the performance of the model with respect to ITA value. 

Although we have not found any evidence of model performance bias under the influence of dataset bias in this particular study, further investigation is needed on datasets with more comprehensive representation. 

Limitations of this work include the following. 1) ITA is an imperfect estimator of skin tone, therefore the measured distributions may contain some degree of error. 2) Ground truth segmentation masks were lacking in both datasets, requiring automatic annotation on most images, which may have introduced additional error. 3) Studies were carried out on the training dataset for ISIC2018 only. The challenge's held-out test dataset contains images drawn from additional statistical distributions (other hospitals and geographical locations), suggesting that test set performance may be lower than observed validation performance on the training dataset. 

\bibliographystyle{unsrt}
\bibliography{bibliography.bib}

\begin{thebibliography}{10}

\bibitem{rotemberg2019role}
Veronica Rotemberg, Allan Halpern, Stephen~W. Dusza, and Noel C.~F. Codella.
\newblock The role of public challenges and data sets towards algorithm
  development, trust, and use in clinical practice.
\newblock {\em Seminars in Cutaneous Medicine and Surgery}, 38(1):E38--E42,
  March 2019.

\bibitem{barocas2016big}
Solon Barocas and Andrew~D. Selbst.
\newblock Big data's disparate impact.
\newblock {\em California Law Review}, 104(3):671--732, June 2016.

\bibitem{bellamy}
Rachel K.~E. {Bellamy}, Kuntal {Dey}, Michael {Hind}, Samuel~C. {Hoffman},
  Stephanie {Houde}, Kalapriya {Kannan}, Pranay {Lohia}, Sameep {Mehta},
  Aleksandra {Mojsilovi\'c}, Seema {Nagar}, Karthikeyan~Natesan {Ramamurthy},
  John {Richards}, Diptikalyan {Saha}, Prasanna {Sattigeri}, Moninder {Singh},
  Kush~R. {Varshney}, and Yunfeng {Zhang}.
\newblock Think your artificial intelligence software is fair? {T}hink again.
\newblock {\em IEEE Software}, 36(4):76--80, July 2019.

\bibitem{karimkhani2017global}
Chante Karimkhani, Robert~P. Dellavalle, Luc~E. Coffeng, Carsten Flohr,
  Roderick~J. Hay, Sin{\'e}ad~M. Langan, Elaine~O. Nsoesie, Alize~J. Ferrari,
  Holly~E. Erskine, Jonathan~I. Silverberg, Theo Vos, and Mohsen Naghavi.
\newblock Global skin disease morbidity and mortality: An update from the
  global burden of disease study 2013.
\newblock {\em JAMA Dermatology}, 153(5):406--412, May 2017.

\bibitem{bray2018global}
Freddie Bray, Jacques Ferlay, Isabelle Soerjomataram, Rebecca~L. Siegel,
  Lindsey~A. Torre, and Ahmedin Jemal.
\newblock Global cancer statistics 2018: {GLOBOCAN} estimates of incidence and
  mortality worldwide for 36 cancers in 185 countries.
\newblock {\em CA: A Cancer Journal for Clinicians}, 68(6):394--424,
  November/December 2018.

\bibitem{cdc2014}
{Centers for Disease Control and Prevention}.
\newblock {US skin cancer costs rise from 2002 through 2011}.
\newblock Online, November 2014.
\newblock Available:
  \url{https://www.cdc.gov/media/releases/2014/p1110-skin-cancer.html}.

\bibitem{marchetti2015screening}
Michael~A. Marchetti, Esther Chung, and Allan~C. Halpern.
\newblock Screening for acral lentiginous melanoma in dark-skinned individuals.
\newblock {\em JAMA Dermatology}, 151(10):1055--1056, October 2015.

\bibitem{wu2011racial}
Xiao-Cheng Wu, Melody~J. Eide, Jessica King, Mona Saraiya, Youjie Huang,
  Charles Wiggins, Jill~S. Barnholtz-Sloan, Nicolle Martin, Vilma Cokkinides,
  Jacqueline Miller, Pragna Patel, Donatus~U. Ekwueme, and Julian Kim.
\newblock Racial and ethnic variations in incidence and survival of cutaneous
  melanoma in the {U}nited {S}tates, 1999-2006.
\newblock {\em Journal of the American Academy of Dermatology},
  65(5):S26.e1--S26.e13, November 2011.

\bibitem{mahendraraj2017}
Krishnaraj Mahendraraj, Komal Sidhu, Christine S.~M. Lau, Georgia~J. McRoy,
  Ronald~S. Chamberlain, and Franz~O. Smith.
\newblock Malignant melanoma in {A}frican–{A}mericans: A population-based
  clinical outcomes study involving 1106 {A}frican–{A}merican patients from
  the surveillance, epidemiology, and end result ({SEER}) database
  (1988–-2011).
\newblock {\em Medicine}, 96(15):e6258, April 2017.

\bibitem{gohara2015skin}
M.~Gohara.
\newblock Skin cancer: An {A}frican perspective.
\newblock {\em British Journal of Dermatology}, 173(Suppl. 2):17--21, July
  2015.

\bibitem{kundu2013dermatologic}
Roopal~V. Kundu and Stavonnie Patterson.
\newblock Dermatologic conditions in skin of color: Part i. special
  considerations for common skin disorders.
\newblock {\em American Family Physician}, 87(12):850--856, June 2013.

\bibitem{STOECKER1992145}
William~V. Stoecker and Randy~H. Moss.
\newblock Editorial: Digital imaging in dermatology.
\newblock {\em Computerized Medical Imaging and Graphics}, 16(3):145--150,
  May--June 1992.

\bibitem{KOROTKOV201269}
Konstantin Korotkov and Rafael Garcia.
\newblock Computerized analysis of pigmented skin lesions: A review.
\newblock {\em Artificial Intelligence in Medicine}, 56(2):69--90, October
  2012.

\bibitem{ibmbookchapter}
Mani Abedini, Qiang Chen, Noel Codella, Rahil Garnavi, Xingzhi Sun, M.~Emre
  Celebi, Teresa Mendonca, and Jorge~S. Marques.
\newblock Accurate and scalable system for automatic detection of malignant
  melanoma.
\newblock In M.~Emre Celebi, Teresa Mendonca, and Jorge~S. Marques, editors,
  {\em Dermoscopy Image Analysis}. CRC Press, 2015.

\bibitem{8627921}
M.~Emre Celebi, Noel Codella, and Alan Halpern.
\newblock Dermoscopy image analysis: Overview and future directions.
\newblock {\em IEEE Journal of Biomedical and Health Informatics},
  23(2):474--478, March 2019.

\bibitem{8660594}
M.~Emre Celebi, Noel Codella, Alan Halpern, and Dinggang Shen.
\newblock Guest editorial: Skin lesion image analysis for melanoma detection.
\newblock {\em IEEE Journal of Biomedical and Health Informatics},
  23(2):479--480, March 2019.

\bibitem{Codella2016}
Noel C.~F. Codella, Quoc-Bao Nguyen, Sharath Pankanti, David~A. Gutman, Brian
  Helba, Allan~C. Halpern, and John~R. Smith.
\newblock Deep learning ensembles for melanoma recognition in dermoscopy
  images.
\newblock {\em IBM Journal of Research and Development}, 61(4/5):5,
  July/September 2016.

\bibitem{haenssle2018man}
H.~A. Haenssle, C.~Fink, R.~Schneiderbauer, F.~Toberer, T.~Buhl, A.~Blum,
  A.~Kalloo, A.~Ben Hadj~Hassen, L.~Thomas, A.~Enk, and L.~Uhlmann.
\newblock Man against machine: Diagnostic performance of a deep learning
  convolutional neural network for dermoscopic melanoma recognition in
  comparison to 58 dermatologists.
\newblock {\em Annals of Oncology}, 29(8):1836--1842, August 2018.

\bibitem{naturepaper}
Andre Esteva, Brett Kuprel, Roberto~A. Novoa, Justin Ko, Susan~M. Swetter,
  Helen~M. Blau, and Sebastian Thrun.
\newblock Dermatologist-level classification of skin cancer with deep neural
  networks.
\newblock {\em Nature}, 542(7639):115--118, February 2017.

\bibitem{adamson2018machine}
Adewole~S. Adamson and Avery Smith.
\newblock Machine learning and health care disparities in dermatology.
\newblock {\em JAMA Dermatology}, 154(11):1247--1248, November 2018.

\bibitem{Merler2019}
Michele Merler, Nalini Ratha, Rogerio~S. Feris, and John~R. Smith.
\newblock Diversity in faces.
\newblock arXiv:1901.10436, April 2019.

\bibitem{ISIC2018}
{International Skin Imaging Collaboration}.
\newblock {ISIC} 2018: Skin lesion analysis towards melanoma detection, 2018.
\newblock Available: \url{https://challenge2018.isic-archive.com/}.

\bibitem{Codella2019}
Noel Codella, Veronica Rotemberg, Philipp Tschandl, M.~Emre Celebi, Stephen
  Dusza, David Gutman, Brian Helba, Aadi Kalloo, Konstantinos Liopyris, Michael
  Marchetti, Harald Kittler, and Allan Halpern.
\newblock Skin lesion analysis toward melanoma detection 2018: A challenge
  hosted by the international skin imaging collaboration {(ISIC)}.
\newblock arXiv:1902.03368, March 2019.

\bibitem{Tschandl2018}
Philipp Tschandl, Cliff Rosendahl, and Harald Kittler.
\newblock Data descriptor: The {HAM10000} dataset, a large collection of
  multi-source dermatoscopic images of common pigmented skin lesions.
\newblock {\em Scientific Data}, 5:180161, August 2018.

\bibitem{chaturvedi2019skin}
Saket~S. Chaturvedi, Kajol Gupta, and Prakash Prasad.
\newblock Skin lesion analyser: An efficient seven-way multi-class skin cancer
  classification using {MobileNet}.
\newblock arXiv:1907.03220, August 2019.

\bibitem{buolamwini2018gender}
Joy Buolamwini and Timnit Gebru.
\newblock Gender shades: Intersectional accuracy disparities in commercial
  gender classification.
\newblock In {\em Proceedings of the Conference on Fairness, Accountability and
  Transparency}, pages 77--91, February 2018.

\bibitem{raji2019actionable}
Inioluwa~Deborah Raji and Joy Buolamwini.
\newblock Actionable auditing: Investigating the impact of publicly naming
  biased performance results of commercial {AI} products.
\newblock In {\em Proceedings of the AAAI/ACM Conference on AI, Ethics, and
  Society}, pages 429--435, January 2019.

\bibitem{muthukumar2019color}
Vidya Muthukumar.
\newblock Color-theoretic experiments to understand unequal gender
  classification accuracy from face images.
\newblock In {\em Proceedings of the IEEE Conference on Computer Vision and
  Pattern Recognition Workshops}, June 2019.

\bibitem{wilson2019predictive}
Benjamin Wilson, Judy Hoffman, and Jamie Morgenstern.
\newblock Predictive inequity in object detection.
\newblock arXiv:1902.11097, February 2019.

\bibitem{sun2016benchmark}
Xiaoxiao Sun, Jufeng Yang, Ming Sun, and Kai Wang.
\newblock A benchmark for automatic visual classification of clinical skin
  disease images.
\newblock In {\em Proceedings of the European Conference on Computer Vision},
  pages 206--222, October 2016.

\bibitem{he2017mask}
Kaiming He, Georgia Gkioxari, Piotr Doll{\'a}r, and Ross~B. Girshick.
\newblock Mask {R-CNN}.
\newblock arXiv:1703.06870, January 2018.

\bibitem{johnson2018adapting}
Jeremiah~W. Johnson.
\newblock Automatic nucleus segmentation with {Mask-RCNN}.
\newblock In {\em Proceedings of the Computer Vision Conference}, pages
  399--407, April 2019.

\bibitem{EilersBGBGNKR2013}
Steven Eilers, Daniel~Q. Bach, Rikki Gaber, Hanz Blatt, Yanina Guevara, Katie
  Nitsche, Roopal~V. Kundu, and June~K. Robinson.
\newblock Accuracy of self-report in assessing {F}itzpatrick skin phototypes
  {I} through {VI}.
\newblock {\em JAMA Dermatology}, 149(11):1289--1294, November 2013.

\bibitem{WilkesWPR2015}
Marcus Wilkes, Caradee~Y. Wright, Johan~L. du~Plessis, and Anthony Reeder.
\newblock {F}itzpatrick skin type, individual typology angle, and melanin index
  in an {A}frican population.
\newblock {\em JAMA Dermatology}, 151(8):902--903, August 2015.

\bibitem{casale2015extreme}
Giuseppe~R. Casale, Anna~Maria Siani, Henri Di{\'e}moz, Giovanni Agnesod,
  Alfio~V. Parisi, and Alfredo Colosimo.
\newblock Extreme {UV} index and solar exposures at {P}lateau {R}osà (3500 m
  a.s.l.) in {V}alle {d'A}osta {R}egion, {I}taly.
\newblock {\em Science of the Total Environment}, 512--513:622--630, April
  2015.

\end{thebibliography}

\end{document}